\documentclass[10pt,letterpaper]{article}
\usepackage[top=0.85in,left=2.75in,footskip=0.75in]{geometry}

\usepackage{amsmath,amssymb}
\usepackage{bm}
\usepackage{changepage}

\usepackage[utf8x]{inputenc}

\usepackage{textcomp,marvosym}

\usepackage{cite}

\usepackage{nameref,hyperref}

\usepackage{microtype}
\DisableLigatures[f]{encoding = *, family = * }

\usepackage[table]{xcolor}

\usepackage{array}

\newcolumntype{+}{!{\vrule width 2pt}}

\newlength\savedwidth

\newcommand\thickhline{\noalign{\global\savedwidth\arrayrulewidth\global\arrayrulewidth 2pt}%
\hline
\noalign{\global\arrayrulewidth\savedwidth}}

\raggedright
\usepackage[aboveskip=1pt,labelfont=bf,labelsep=period,justification=raggedright,singlelinecheck=off]{caption}

\makeatletter
\renewcommand{\@biblabel}[1]{\quad#1.}
\makeatother

\usepackage{lastpage,fancyhdr,graphicx}
\usepackage{epstopdf}
\fancyheadoffset[L]{2.25in}
\fancyfootoffset[L]{2.25in}
\begin{document}
\vspace*{0.2in}

\begin{flushleft}
{\Large
\textbf\newline{Evaluation of soccer team defense based on prediction models of ball recovery and being attacked: A pilot study} 
}
\newline
\\
Kosuke Toda\textsuperscript{1},
Masakiyo Teranishi\textsuperscript{2},
Keisuke Kushiro\textsuperscript{1},
Keisuke Fujii\textsuperscript{2,3,4 *}
\\
\bigskip
\textbf{1} Graduate School of Human and Environmental Studies, Kyoto University, Kyoto, Kyoto, Japan
\\
\textbf{2} Graduate School of Informatics, Nagoya University, Nagoya, Aichi, Japan.
\\
\textbf{3} RIKEN Center for Advanced Intelligence Project, Fukuoka, Fukuoka, Japan.

\textbf{4} PRESTO, Japan Science and Technology Agency, Kawaguchi, Saitama, Japan.
\\
\bigskip

%
%





* fujii@i.nagoya-u.ac.jp

\end{flushleft}
\section*{Abstract}
With the development of measurement technology, data on the movements of actual games in various sports can be obtained and used for planning and evaluating the tactics and strategy. Defense in team sports is generally difficult to be evaluated because of the lack of statistical data. Conventional evaluation methods based on predictions of scores are considered unreliable because they predict rare events throughout the game. Besides, it is difficult to evaluate various plays leading up to a score.

In this study, we propose a method to evaluate team defense from a comprehensive perspective related to team performance by predicting ball recovery and being attacked, which occur more frequently than goals, using player actions and positional data of all players and the ball.  
Using data from 45 soccer matches, we examined the relationship between the proposed index and team performance in actual matches and throughout a season.

Results show that the proposed classifiers predicted the true events (mean F1 score $>$ 0.483) better than the existing classifiers which were based on rare events or goals (mean F1 score $<$ 0.201). 
Also, the proposed index had a moderate correlation with the long-term outcomes of the season ($r =$ 0.397). These results suggest that the proposed index might be a more reliable indicator rather than winning or losing with the inclusion of accidental factors. 


\section*{Introduction}
The development of measurement technology has allowed for the generation of data on the movements in various sports games for use in planning and evaluating the tactics and strategy. For example, tracking data during a game of soccer, including the positional data of the players and ball, is commonly used for players' conditioning (e.g., running distance or the number of sprints) \cite{andrzejewski2013analysis,andrzejewski2015sprinting}.
However, during a soccer match, all 22 players and the ball interact in complex ways for scoring goals or preventing being scored (it is sometimes referred to as conceding) for each team.
Hence, it is then necessary to evaluate the performance of not only individuals but also the entire team \cite{fujii2021data}.In particular, defensive tactics are considered difficult to evaluate because of the limited amount of available statistics, such as goals scored in the case of attacks.

There are three main approaches to quantitatively evaluate teams and players in soccer, mainly from an attacking perspective. 
\textcolor{black}{Note that the evaluation of defense is almost the same as that of offense, because the defenders try to prevent being scored and the attackers try to score.}
The first approach is based on scoring prediction, which evaluates plays based on changes in the expected values of goals scored and conceded based on a prediction of scoring using tracking data \cite{mchale2007modelling,mchale2012development,pappalardo2019playerank,decroos2019player} (reviewed in \cite{vanroy2020valuing})
and action data such as dribbling and passing \cite{Decroos19}, as well as other rule-based methods (e.g., based on the two distances of the ball player from the nearest defender and the goal \cite{Teranishi20}).
\textcolor{black}{In particular, Valuing Actions by Estimating Probabilities (VAEP) \cite{Decroos19} was recently proposed as a framework for valuing player actions in soccer, which is based on the prediction model of the scores and concedes utilizing on-the-ball action such as dribbles and passes. }
The second approach is used to evaluate plays such as passes and effective attacks which lead to shots. For example, a previous study evaluated the value of passes based on relationships to the expected score and the difficulty in successfully completing a pass \cite{Power17}. 
An effective attack can be defined as a play that will likely lead to a score\cite{Ueda14}.
Previous studies have analyzed pass networks \cite{Yamamoto11} and three player interactions \cite{Yokoyama11,Yokoyama18}, pass reception \cite{Fujii20cognition}, and of the related defensive weaknesses \cite{Llana20}. For defense, researchers have evaluated interception \cite{Piersma20} and the effectiveness of defensive play by the expected value of a goal-scoring opportunity conceded \cite{Robberechts19}.
For the third approach, a spatial positioning of the players is evaluated by calculating the dominant region with the use of a Voronoi diagram \cite{Taki00} and the Gaussian distribution \cite{Kijima14}. Recent research has also been conducted on the evaluation of movements that create space for teammates \cite{spearman2018beyond,Fernandez18}.
Other approaches such as self-organizing maps using artificial neural networks have been reviewed in \cite{fujii2021data}.
However, these approaches sometimes have several limitations. For evaluation based on the prediction of scoring (i.e., the first approach), the evaluation is not reliable because it predicts events that are rare throughout a game, and the process leading to the goals is sometimes difficult to evaluate. Furthermore, the second approach to evaluate specific plays that lead to goals and the third approach regarding positioning sometimes ignored the relationship with overall performance such as wins and losses (note that there have been some studies to investigate the relationship between performance indicators and the outcome such as the promotion to the elite leagues in \cite{jamil2021investigation}. Also, since many studies on the first and second approaches have used only the actions and coordinates of players around the ball, it would be difficult to evaluate players at greater distances from the ball and the team as a whole.

To address these issues, we propose a method called \textit{Valuing Defense by Estimating Probabilities} (VDEP), which utilizes the actions and positional data of all players and the ball.
The ultimate goal of defense in soccer is to prevent the opposing team from scoring a goal. However, since goal-scoring scenes are rare events, it may lead to ineffective training of a classifier and evaluating the events in an unreliable manner \textcolor{black}{when the dataset size is limited} (the verification results of the VAEP method \cite{Decroos19} will be presented later). Therefore, to reasonably evaluate the defense of a team, we propose the VDEP method to evaluate important factors for preventing goals from being scored.
The VDEP method evaluates the potential increase in the number of ball recoveries and the potential decrease in the number of effective attacks. The number of effective attacks was chosen instead of the number of shots because \textcolor{black}{we also regard} the scenarios as defensive failures, in which an attacker selects to pass the ball rather than to shoot \textcolor{black}{in the penalty area}. Therefore, in this study, we evaluate the process of defense based on the expected value computed by the classifiers to predict ball recovery and being attacked in an analogous way of the VAEP method \cite{Decroos19} based on the prediction of scoring and conceding.
\textcolor{black}{We can meaningfully discuss the results of VAEP and our approach because of the similar approaches.}

The main contributions of this work are as follows: (i) the proposed method is based on the prediction of ball possession and effective attacks, which occur more frequently than the rare goals; (ii) based on a comprehensive perspective related to team outcomes, we evaluated the team's defense.
Methodologically, we modified the existing method called VAEP  \cite{Decroos19}, which is based on the classifiers to predict scoring and conceding, so that the defensive process can be evaluated by applying the approach to ball recovery and being attacked. We validated the classifiers of the proposed and existing methods and shows that the proposed classifiers predicted the true events better than the existing classifiers. Moreover, we examined the relationship between VDEP and the team performance in actual matches and throughout the season, as compared with VAEP. We also presented examples of evaluating a game and a complete season of a specific team.

\section*{Materials and methods}
\subsection*{Dataset}
In this study, we used event data (i.e., labels of actions, such as passing and shooting, recorded at 30 Hz and the simultaneous xy coordinates of the ball) and tracking data (i.e., xy coordinates of all players and the ball recorded at 25 Hz) with 45 games from week 30 to week 34 of the Meiji Yasuda Seimei J1 League 2019 season.
\textcolor{black}{Note that we used ``event'' as the above meaning based on the previous studies \cite{Decroos19,pappalardo2019playerank}.}
These data were provided by Research Center for Medical and
Health Data Science in the Institute of Statistical Mathematics \textcolor{black}{(an academic organization)} and Data Stadium Inc.
That is, we did not estimate xy coordinates and directly used it for the subsequent analysis. 
Data acquisition was based on the contract between the soccer league (J League) and the company (Data Stadium, Inc), not between the players and us.
The company was licensed to acquire this data and sell it to third parties, and it was guaranteed that the use of the data would not infringe on any rights of J League players or teams.
We obtained the data by participating in a competition hosted by the 
\textcolor{black}{academic organizations and the central idea of this study was independent of the competition (not restricted by the competition).} 

In all 45 games, there were 106 goals scored, 1,174 shots, 3,701 effective attacks, and 9,408 ball recoveries (all based on the provided event data).
An effective attack is defined as an event that finally ends in a shot or penetrates the penalty area. Also, ball recovery is defined as a change in the attacking team before or after the play due to some factors other than an effective attack.
In this study, an effective attack is defined as \textit{being attacked} from the defender's perspectives.
\textcolor{black}{It should be noted that for the labeling of effective attack/ball recovery, we labeled each event (not an attack segment) and there are four combinations of positive/negative effective attack/ball recovery.
Since in soccer the attacking team transitions sometimes occur in a quite short time, we did not explicitly define an attack segment based on the previous work \cite{Decroos19}.}

\subsection*{Proposed Method}
Based on the motivation mentioned in the Introduction, here we describe the details of our proposed method.
Suppose that the state of the game is given by $S = [s_1,\ldots,s_N]$ in chronological order.
We consider $s_i = [a_i,o_i]$, whereas the previous study \cite{Decroos19,Robberechts19} used only $a_i$, which includes the $i$th action involving the ball and its coordinates.
The proposed method utilizes classifiers trained with the state $s_i$, which includes the feature $o_i$ far from the ball (off-ball) at the time of the action.
Since all defensive and offensive actions in this study are evaluated from the defender's point of view, the following time index $i$ is used as the $i$th \textit{event}.

Given the game state $S_i$ of a certain interval, we define the probability of future ball recovery $P_{recoveries}(S_i)$ and the probability of being attacked $P_{attacked}(S_i)$ in a state $S_i$ at an event $i$ based on the classifier trained from the data.
Defensive players are considered to act so that $P_{recoveries}(S_i)$ becomes higher or $P_{attacked}(S_i)$ becomes lower.
Therefore, the value of defense in the proposed method $V_{vdep}$ is defined parameter $C$ that adjusts the values of ball recoveries and effective attacks as follows:
\begin{equation}
\label{eq:vvdep}
    V_{vdep}(S_i) = P_{recoveries}(S_i) -C*P_{attacked}(S_i). 
\end{equation}
In this study, we adjusted these values based on the frequency of each event in the training data. As described below, we determined $C\approx3.9$ because the ratio of ball recoveries and effective attacks is approximately $9,408:3,701 \approx 3.9:1$ (the value differs for each of five-fold cross-validation).
In this study, we assume that the importance of ball recoveries and effective attacks is determined based on these frequencies, but this may be controversial. We discuss this point in the Discussion section as our future work.

Since the main aim of this study is to evaluate the team, we define the evaluation value per game for team $p$ as follows:
\begin{equation}
\label{eq:rvdep}
    R_{vdep}(p) = \frac{1}{M}\Sigma_{S_i\in \bm{S}_{M}^p} V_{vdep}(S_i),
\end{equation}
where $M$ is the number of events for team $p$ in a match and $\bm{S}_M^p$ is the set of states $S$ of team $p$ up to the $M$th event. Similarly, the mean values of only $P_{recoveries}$ and $P_{attacked}$ are defined as $R_{recoveries}(p)$ and $R_{attacked}(p)$, respectively. For the VAEP\cite{Decroos19} method in the previous study, the value averaged by the playing time of each player was used.
However, since the time each team played the game was almost the same, in this study, each team is evaluated by the sum of $S_{vaep}(p)$ as the VAEP value. Also, $S_{scores}(p)$ and $S_{concedes}(p)$ are used in the analysis as separate evaluation values based on the probabilities of scores $P_{scores}$ and concedes $P_{concedes}$ (note that the VAEP\cite{Decroos19} value is calculated based on the prediction of goals scored and conceded).

\begin{figure}
\centering
\includegraphics[scale=0.6]{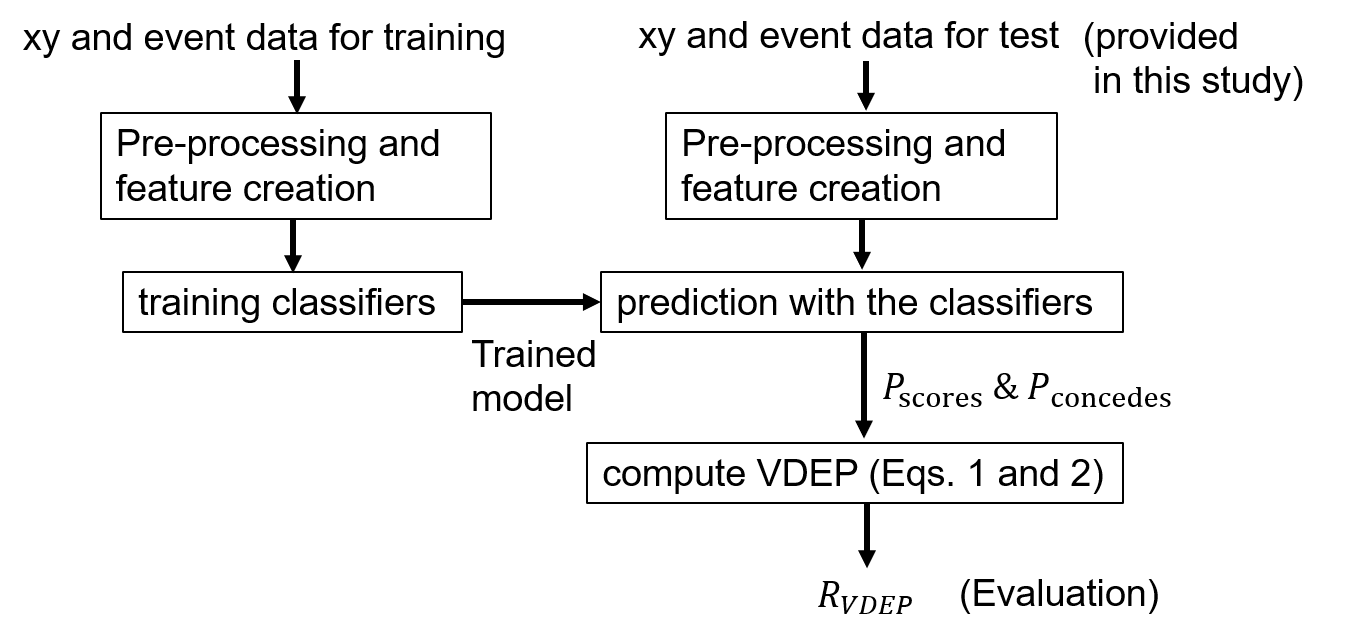}
\caption{\label{fig:diagram}{\bf The flow diagram of the analysis.} \textcolor{black}{The computation of our VDEP is composed of four steps: pre-processing and feature creation, training classifiers, prediction with the classifiers, and computing VDEP.
First, we perform data pre-processing and feature creation for the classifier using provided xy coordinates and event data.
Second, we train the classifiers of ball recoveries and being attacked using xy coordinates in the training dataset. Third, the trained classifiers predict the ball recoveries and being attacked using xy coordinates in the test dataset.} Finally, we computed VDEP using Equations \ref{eq:vvdep} and \ref{eq:rvdep}.
The existing VEAP \cite{Decroos19} can be computed in a similar procedure.
}
\end{figure}

\begin{figure}
\centering
\includegraphics[scale=0.5]{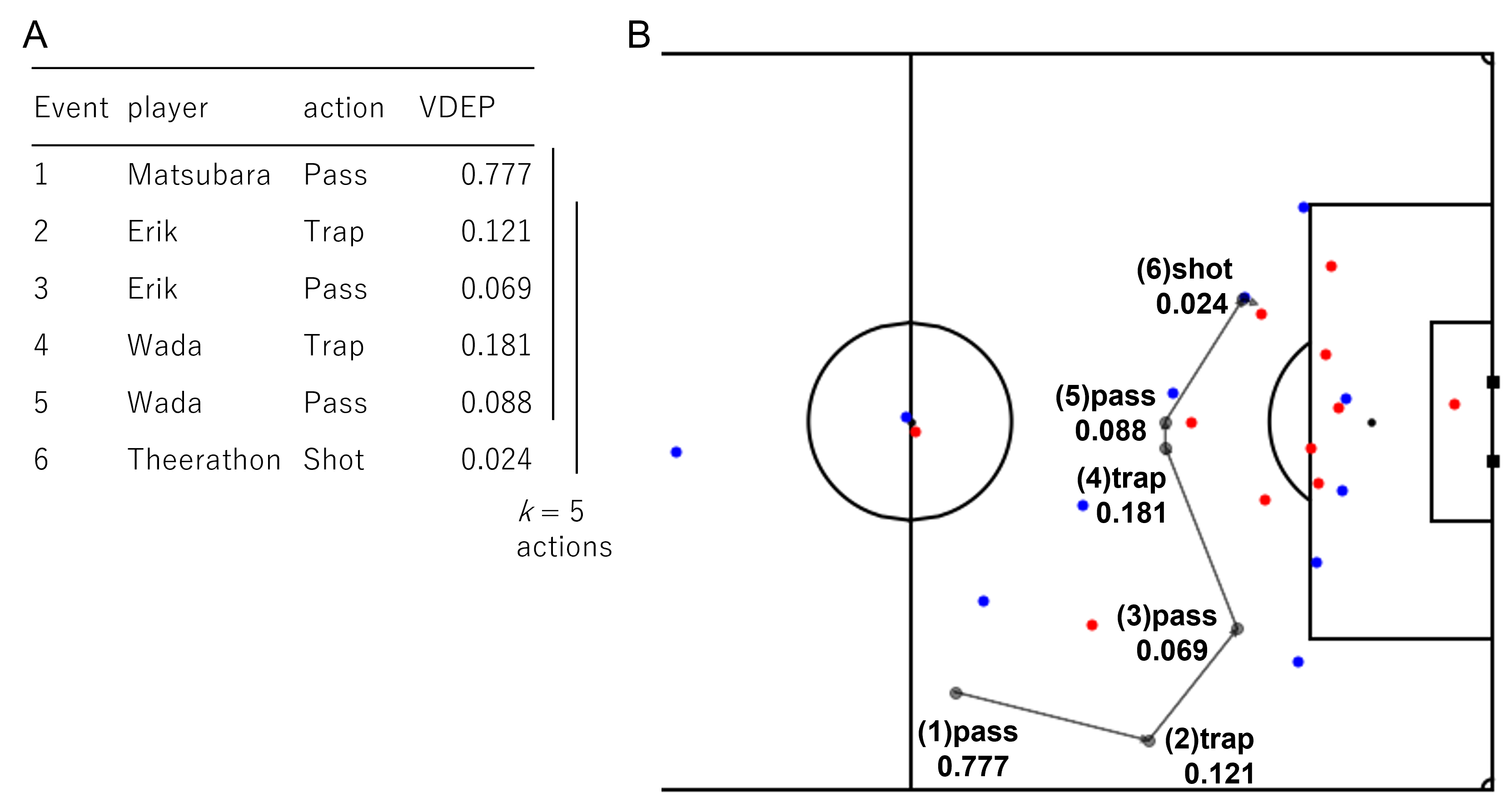}
\caption{{\bf Example of defensive play analysis.} 
(A) The VDEP value for each event is indicated, including the type of event, and the player who took the action. \textcolor{black}{Our classifiers for VDEP predict a ball recovery or an effective attack in the subsequent $k$ actions or events (we set $k=5$ and in this case, all labels are negative).} (B) The position of all players when the shot was performed are visualized (red: defending team; blue: attacking team) and the flow of the event with the ball. In this scene, the VDEP values were positive in all events, suggesting that the defense was not so bad that the goal was conceded. The VDEP values decreased between Matsubara's pass and Erik's trap, and between Erik and Wada's trap and pass, suggesting that the goal was conceded because of a forward pass or because the ball holder was allowed to go free. }
\label{fig:example}
\end{figure}

\textcolor{black}{\subsection*{Pre-processing and Feature Creation}
The flow diagram of the analysis procedure is shown in Fig \ref{fig:diagram}. 
In summary, we perform data pre-processing and feature creation, training classifiers, prediction with the classifiers, and computing VDEP.
In this subsection, we describe pre-processing and feature creation.}
The time range of the input $S_i$ to the classifier was $i$th, $i-1$th, and $i-2$th events in the previous study \cite{Decroos19}.
In this study, since the effect of $s_{i-2}$ on the prediction performance was small in the preliminary experiments, we used $s = [a,o]$ including the $i$th and $i-1$th events.
In the first classification for estimating $P_{recoveries}(S_i)$, we assigned a positive label ($= 1$) to the game state $S_i$ if the defending team in the state $S_i$ recovered the ball in the subsequent $k$ events, and a negative label ($= 0$) if the ball was not recovered. Similarly, in the second classification for estimating $P_{attacked}(S_i)$, we assigned a positive label ($= 1$) to the game state $S_i$ when an effective attack was made in the subsequent $k$ events.
An illustrative example is shown Fig \ref{fig:example}A.
In both classifications, $k$ is a parameter freely determined by the user. If $k$ is small, the prediction is short-term\textcolor{black}{, smaller positive labels, reliable, and obtains unambiguous interpretation, and if $k$ is large, the prediction is long-term, larger positive labels, includes many factors, and obtains ambiguous interpretation. 
Since it is intrinsically difficult to solve this trade-off, we set $k=5$ by the domain knowledge. }

\textcolor{black}{Next, we describe the feature vector creation.
we first created the 36 dimensional features of $i$th and $i-1$th events.}
The feature $a_i$ near the ball in this study was constructed using the event and tracking data. 
Specifically, we used the types of events used in the previous study \cite{Decroos19} (19 types: pass, cross, throw in, free kick, corner kick, trap, foul, tackle, interception, shot, PK, own goal, goalkeeper hand clear, goalkeeper catch, clearance, block, dribble, off-side, and goal kick. for details, see S1 Text). We also used 
\textcolor{black}{the event ID (1 dim), the start/end time and the duration of the event (3 dim.), the xy coordinates of the ball at the start and the end (4 dim.), the displacements of the ball from the start to the end (x, y, the Euclidean norm: 3 dim.), and those from the previous event start (3 dim.), the distance and angle between the ball and the goal (2 dim.), and whether there was a change in offense or defense from the previous event (1 dim.) 
Moreover, in this study, the off-ball feature $o_i$ at the time that the event occurred was included in the state $s_i$. Specifically, we used the x and y coordinates of positions of all players (22 players xy coordinates) and the distance of each player from the ball (22 players), sorted in the order of closest to the ball.
Finally, to reflect the opponent’s attacking ability, we add the opponent team season scores (1 dim.) to the feature.
Therefore, the feature has 139 dimensions in total ($36\times 2 + 22\times 3 + 1 = 139$).}

In the data used in this study, defined by $k=5$ above, the total number of events for all teams was 97,335, with 35,286 positive cases of ball recovery and 13,353 positive cases of being attacked.
In terms of goals scored and conceded for the calculation of the VAEP value \cite{Decroos19}, there were 753 positive cases of goals scored and 227 positive cases of goals conceded (the total number of events was the same, but we set $k=10$ in accordance with the previous study \cite{Decroos19}). These indicate that goals scored and conceded are rare events compared to ball recoveries and being attacked \textcolor{black}{(from the above reasons, it may be meaningful to compare VAEP with $k=10$ and our VDEP with $k=5$). }
Therefore, the goals scored and conceded may not be correctly verified by the area under the receiver operating characteristic curve (AUC) and accuracy \textcolor{black}{in this study with a smaller dataset (compared with the larger dataset in the previous work \cite{Decroos19})} as described below.

\subsection*{\textcolor{black}{Prediction Model Implementation}}
We adopted XGBoost (eXtreme Gradient Boosting) \cite{Chen16}, which was used in the previous study \cite{Decroos19}, as the classifiers to predict ball recoveries and being attacked. 
\textcolor{black}{Gradient tree boosting \cite{friedman2001greedy} has been a popular technique leveraging a prediction algorithm that sequentially produces a model in the form of linear combinations of decision trees and performs well on a variety of learning problems with heterogeneous features, noisy data, and complex dependencies.  
XGBoost \cite{Chen16} is a variant of gradient tree boosting model which can be computed in a faster and more scalable way.}
Note that other classifiers can be used in the same framework.
\textcolor{black}{Although the prediction model itself do not consider the time series structure, according to the previous VAEP framework \cite{Decroos19} and as described above, the prediction models reflect the history of the input ($i$th and $i-1$th events) and that of the output (the subsequent $k$ events).}

When calculating VDEP and VAEP values, we used a five-fold cross-validation procedure. 
Here we define the terms of training, validation, and test (datasets). 
We train the machine learning model using the training dataset, 
validate the model performance using the validation dataset (sometimes for determining some hyper-parameters), and finally test the model performance using the test dataset.
The benefit of such procedure is to verify a model which can test the performance using a new test data (not used during training). 
In our case, the validation data was not used and hyperparameters are predetermined as default in Python library ``xgboost'' (version 1.4.1).
\textcolor{black}{We did not use the early stopping method for the training.}
``Cross validation procedure'' we used here is a test using the test dataset in an analogous way of the usual cross-validation to analyze all data even using a small dataset \cite{Fujii20,Fujii17,Fujii18}. 
In the cross-validation, the original sample is randomly partitioned into five equal sized subsamples. 
Of the five subsamples, a single subsample is regarded as the test data for testing the model, and the remaining four subsamples are used as training (and validation) data. 
The cross-validation process is then repeated five times, with each of the five subsamples used exactly once as the test data. 
There seems to be no best practice to determine the number of folds and researchers usually select such as 5 or 10 in advance. 
A higher number of folds indicates that each model is trained on a larger training set and tested on a smaller test fold, which lead to a lower prediction error because the model can use more training data, but it takes more computational time and the smaller test data may inaccurately evaluate the error.
In our case, the dataset size was not so large and we had five-week games data, thus we selected to use five-fold cross-validation.
Actually, we repeated the learning of classifiers using the data of four weeks (36 games) and a prediction using the data of one week (nine games) five times (i.e., data of all five weeks were finally predicted and evaluated) to analyze all games.
In this study, since we prioritized the analysis among teams, we need to utilize the limited data (5 games for each team) and we assume that the performances for each week game were independent. 
\textcolor{black}{For future work, we need to verify our method using different datasets (e.g., obtained in different environments).}

Our all computations were performed using Python (version 3.6.8).
In particular, we customized the published code of the VAEP method \cite{Decroos19} using Python (\url{https://github.com/ML-KULeuven/socceraction}).
We recorded the computational time of our method (VDEP) and the existing method (VAEP) during five-fold cross validation (i.e., 36 games for training and nine games for testing).
The training time was $351.6$ [s] in VDAP and $212.6$ [s] in VAEP (both 36 games $\times$ 5 in total), and the inference time (i.e., testing) was $3.9$ [s] in VDAP and $3.6$ [s] in VAEP (both 9 games $\times$ 5 in total).
\textcolor{black}{For the computation of these indices in practice (if we have the trained model), we can compute them within one second per game regarding the inference. 
}

\subsection*{Evaluation and Statistical Analysis}
To validate the classifier, we used the F1 score in addition to the AUC and Brier scores used in the previous study \cite{Decroos19}. AUC is calculated by plotting the cumulative distribution function of the true positive rate against the false positive rate. 
\textcolor{black}{The true positive rate is defined as the ratio of the sum of true positives and true negatives to the number of true positives.
The false positive rate is defined as the ratio of the sum of false positives and true negatives to the number of false positives.}
AUC indicates 0.5 for random prediction and 1 for perfect prediction.
Brier score is the mean squared error between the predicted probability and the actual outcome, where a smaller value indicates a better prediction. 
However, these evaluations and (more intuitive) accuracy score may not be better when there are extremely more negative than positive cases, as in this and previous studies.
As an intuitive example, the accuracy of being attacked in VDEP and scored in VAEP will be $1-13353/97335\approx0.863$ and $1-753/97335\approx0.992$ when all negative cases are predicted. AUC and Brier score also have similar problems because these indices evaluate large amounts of true negatives.
\textcolor{black}{Since AUC also evaluates false positive rate, the similar problem to accuracy and Brier score can occur when small true positives and larger true negatives. }
In this study, we also used the F1 score to evaluate whether the true positives can be classified without considering the true negatives. The F1 score is expressed as F1score = (2 $\times$ Precision $\times$ Recall) / (Precision + Recall), where the Recall is equal to the true-positive rate, and the Precision is defined as the ratio of the sum of true positives and true negatives to false positives.  In this index, only true positives are evaluated, not true negatives.
To compare F1 scores among the various classifiers for testing our hypothesis (other AUC and Brier scores are shown only as references), since the hypothesis of homogeneity of variances between methods was not rejected with Levene’s test, a one-way analysis of variance was performed. As a post-hoc comparison, Tukey's test was used within the factor where a significant effect in one-way analysis of variance was found. 
Furthermore, the contribution of the input variables to the prediction of the VDEP method was calculated by SHAP (SHapley Additive exPlanations) \cite{Lundberg17}, which utilizes an interpretable approximate model of the original nonlinear prediction model. 

For the evaluation of defense using the VDEP and VAEP values \cite{Decroos19}, we present examples to quantitatively and qualitatively evaluate a game and a season of a specific team.
Next, we examined the relationships with the outcomes of actual games (goals scored, conceded, and winning points, where win, draw, and lose were assigned as 3, 1, and 0 points, respectively) and the relationship with the team results throughout the season using the Pearson's correlation coefficient among all 18 teams. 

For all statistical analysis, $p<0.05$ was considered significant. However, since the sample size was small ($N=18$) in the correlation analysis, the $r$ value indicating the magnitude of the correlation was also used as an effect size for evaluation.
As described in a previous study \cite{guilford1950fundamental},
correlation coefficients less than $0.20$ were interpreted as \textit{slight almost negligible relationships}, between $0.20$ and $0.40$ as \textit{low correlation}; between $0.40$ and $0.70$ as \textit{moderate correlation}; between $0.70$ and $0.90$ as \textit{high correlation} and correlation greater than $0.90$ as \textit{very high correlation}.
In this study, the correlation coefficients were rounded off to the third decimal place for interpretation.
All statistical analyses were performed using SciPy (version 1.5.4) in the Python library.

\section*{Results}
\subsection*{Verification of Classifiers}
To validate the VDEP and VAEP\cite{Decroos19} methods, we first investigated the prediction performances of their classifiers.
As mentioned earlier, there are two classifiers of pass recoveries and being attacked in VDEP (similarly, the existing VAEP has two classifiers of scores and concedes). 
These classifiers predict probabilities of pass recoveries ($P_{recoveries}$) and being attacked ($P_{attacked}$) in VDEP (similarly, probabilities of scores $P_{scores}$ and concedes $P_{concedes}$ in VAEP).
In Table~\ref{table1}, the classifiers of VDEP shows better predictions than those of VAEP\cite{Decroos19} (note that the output and number of occurrences to be predicted are different).
The AUCs of $P_{recoveries}$ and $P_{attacked}$ in VDEP were better than those of $P_{scores}$ and $P_{concedes}$ in VAEP, and vice versa in regard to the Brier scores.
\textcolor{black}{Specifically, Brier score is directly affected by larger true negatives (see the above accuracy example) and thus it may show a better result in VAEP than that of our VDEP. 
On the other hand, AUC evaluates true-positive rate and the biased effect was smaller than Brier score, thus it shows a better result in our VDEP than that of VAEP. }
However, again, these indices may not be validly evaluated because they include a large number of true negatives in the evaluation (thus, we did not perform statistical analysis in these variables).
Instead, the F1 score was calculated, and the statistical analysis identified significant main effect among $P_{recoveries}$, $P_{attacked}$, and $P_{scores}$ ($F=144.40,~ p < 1.0 \times 10^{-6}$; $P_{concedes}$ was eliminated because the average is near zero value). 
The post-hoc analysis shows that F1 scores of VDEP ($P_{recoveries}$, $P_{attacked}$) were significantly higher than that of $P_{scores}$ ($ps < 0.002$).
This indicates that the VDEP method predicted true positives correctly, while the VAEP did not.
For details, the confusion matrices of the four classifiers are shown in \nameref{S1_Fig}.

\begin{table}[!ht]
\begin{adjustwidth}{-2.25in}{0in} 
\centering
\caption{
{\bf Evaluation of classifiers for the proposed and conventional methods.}}
\begin{tabular}{c|c|c|c}
\hline
 & AUC & Brier score & F1 score \\ \thickhline
$P_{recoveries}$ & 0.770 $\pm$ 0.014& 0.184 $\pm$ 0.009& 0.522 $\pm$ 0.036\\$P_{attacked}$ & 0.862 $\pm$ 0.003& 0.079 $\pm$ 0.003& 0.484 $\pm$ 0.038\\ \hline
$P_{scores}$\cite{Decroos19} & 0.698 $\pm$ 0.066& 0.007 $\pm$ 0.002& 0.201 $\pm$ 0.021\\
$P_{concedes}$\cite{Decroos19} & 0.701 $\pm$ 0.040& 0.003 $\pm$ 0.001& 0.000 $\pm$ 0.000\\\hline
\end{tabular}
\begin{flushleft} 
\end{flushleft}
\label{table1}
\end{adjustwidth}
\end{table}

\begin{figure}
\centering
\includegraphics[scale=0.4]{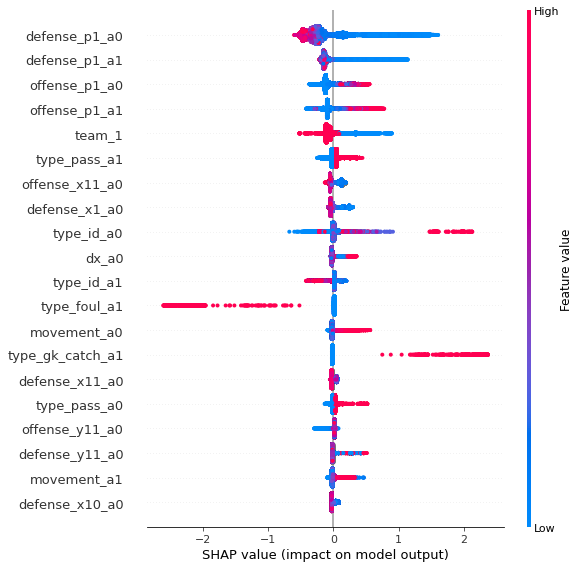}
\caption{\label{fig:shap_recover}{\bf Contribution of the input variables to the prediction of $P_{recoveries}$. } The input variables related to the prediction of $P_{recoveries}$ are presented in the order of their contributions. Of the top 20 features, those at the top had greater contribution than those at the bottom. Each dot represents each event. The color represents the value of the feature (blue and red indicate low and high, respectively). The horizontal axis shows the impact on the prediction (strongly positive and negative impacts are plotted to the right and left, respectively). For example, when the value of type\_foul\_a1 is 1, the prediction is likely to be zero.
\textcolor{black}{For variable names, the a\_0,1 means an analyzed event or one previous event. The x,y,p mean x,y coordinates and distance from the ball. The team\_1 shows whether there was a change in offense or defense from the one previous event. The dx and movement are the x displacements and the Euclidean norm of the ball from the start to the end. The type\_ is the  variables related to the events.}}
\end{figure}

\begin{figure}
\centering
\includegraphics[scale=0.4]{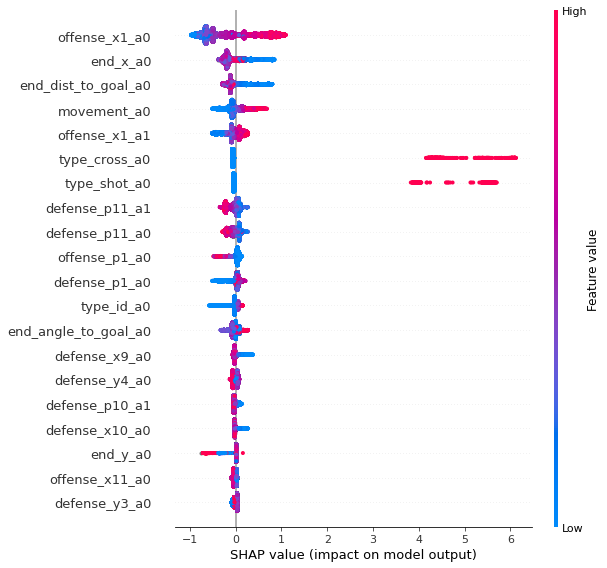}
\caption{\label{fig:shap_attacked}{\bf Contribution of the input variables to the prediction of $P_{attacked}$.} The input variables related to the prediction of $P_{attacked}$ are shown in the order of their contribution. The configuration is the same as in Fig \ref{fig:shap_recover}. For example, when the value of offense\_x1\_a0 is positively and negatively large, the prediction value is also likely to be positively and negatively large, respectively. 
\textcolor{black}{In addition to the variable names shown in Fig \ref{fig:shap_recover}, the end\_ means the variables at the end of the analyzed events.} }
\end{figure}

Next, the contribution of the input variables to the prediction of the VDEP method was calculated by SHAP \cite{Lundberg17}. For $P_{recoveries}$, in Fig \ref{fig:shap_recover}, the distance to the ball of the defender closest to the ball had the highest contribution, followed by the events where there was an offensive or defensive change immediately beforehand. For $P_{attacked}$, in Fig \ref{fig:shap_attacked}, the x-coordinate of the attacker closest to the ball (in the direction of the goal) and the displacement of the attacker from the beginning to the end of the event had the largest contribution, followed by the distance to the ball of the defender closest to the ball.

\subsection*{Examples of Team Defense Evaluation}

\subsubsection*{Evaluation of a defensive play}
An advantage of the VDEP method is the ability to show the effectiveness of the formation of the defending team against the attacking team at a particular moment in the game. For example, with the use of VDEP for a goal conceded, it can be easily understood where the factor of the goal is placed in the series of events.
As an example, consider the first goal in the match between Yokohama F. Marinos and FC Tokyo shown in Fig \ref{fig:example}. A positive VDEP value can be interpreted as a good defense and a negative value as a bad defense. In this example, the VDEP values were positive in all events, indicating that the defense was not so bad that the goal was conceded. However, to be precise, the VDEP values decreased between Matsubara's pass and Erik's trap, and between Erik and Wada's trap and pass, suggesting that the goal was conceded because of a forward pass or because the ball holder was allowed to go free. 

\subsubsection*{Evaluation of a game}
Since it is sometimes difficult to score goals in soccer, the team that dominates the game does not always win. Therefore, to continuously strengthen the team, it is necessary to analyze the game regardless of the immediate outcomes. 
The VDEP method is expected to be used as a more stable evaluation index than wins and losses which are limited by contingent factors.

For example, in the match between Yokohama F. Marinos and FC Tokyo, Yokohama won the match by a score of 3 to 0. We examined the reasons for the unexpectedly large gap in the matchup of the two top teams (the numbers of shots taken by both teams were the same in the game). 
Although $R_{recoveries}$ for Yokohama (0.371) was better than that for Tokyo (0.348), $R_{attacked}$ and $R_{vdep}$ for Tokyo (0.116 and 0.049) were better than that for Yokohama (0.159 and -0.040).  These indicate that Tokyo's defense made it difficult for Yokohama to score goals. As in this game, there are cases where the evaluation results do not match the game outcome, e.g., when the quality of shots was better even if the defensive evaluation was good (note that the proposed method did not reflect how likely an effective attack is to score). Thus, the use of the VDEP method to quantitatively evaluate the defense of each match will allow for a more detailed analysis than wins and scores.

Statistically, correlation analysis was performed between the outcome of the game and the proposed and existing indices (analyzed data is given in \nameref{S1_Data}).
In the case of $R_{vdep}$, there were moderate positive correlations with winning points ($r_{16} = 0.464, p = 0.050$) and low positive correlation with goals scored ($r_{16} = 0.392, p = 0.106$).
In the case of $S_{vaep}$, there were high positive correlation with winning points ($r_{16} = 0.830, p < 0.001$) and very high positive correlation with goals scored ($r_{16} = 0.953, p < 0.001$).
It is obvious that $S_{vaep}$ can sufficiently predict the number of goals scored in a match because it is based on the prediction of scores.
Interestingly, even though $S_{vaep}$ is also based on the prediction of conceded goals, it had slight almost negligible relationships with goals conceded ($r_{16} = -0.040, p > 0.05$). 
On the other hand , $R_{vdep}$ had low correlation with the goals scored in the game ($r_{16} = -0.245, p > 0.05$).


\subsubsection*{Defensive evaluation of teams in multiple games}
\begin{figure}[!h]
\includegraphics[scale=0.7]{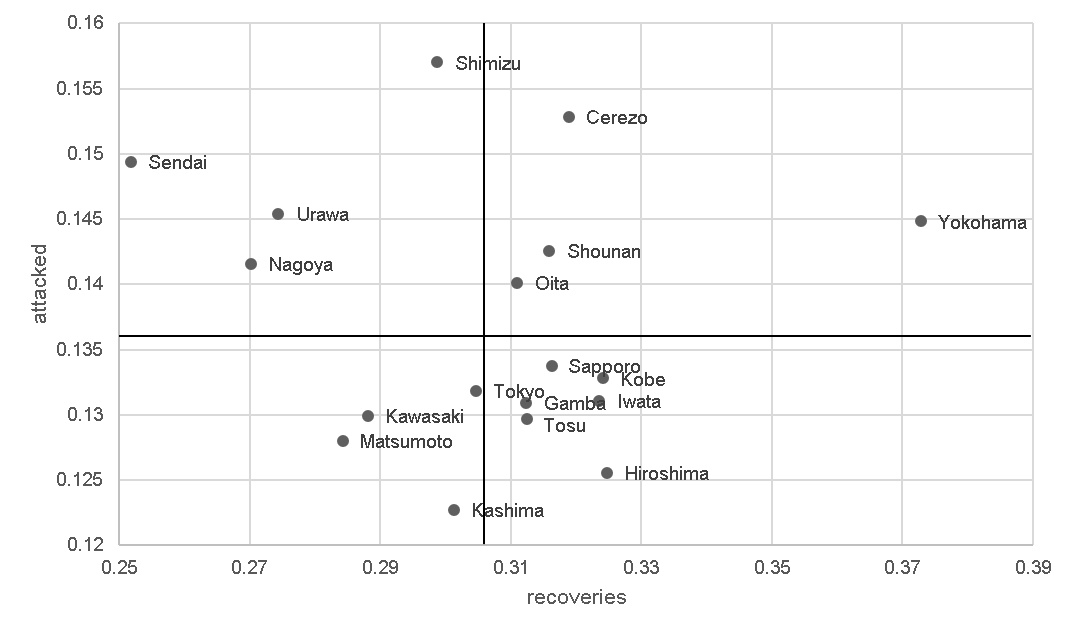}
\caption{{\bf Defensive evaluation of teams in multiple games.}
The vertical axis is $R_{attacked}$ and the horizontal axis is $R_{recoveries}$.
The vertical and horizontal lines are the averaged values of $R_{attacked}$ and $R_{recoveries}$ among all teams, respectively.
The more points plotted to the right, the more likely the defense is to recover the ball, and the more points plotted below, the less likely the defense is to concede. The black line is the league average. For example, Yokohama defended with a high probability of recovering the ball. On the other hand, the probability of being attacked was also high, suggesting that the team adopted a high-risk yet high-return defensive tactic. }
\label{fig5}
\end{figure}

It is also possible to characterize and evaluate team defenses throughout a season using the VDEP values in multiple games.
 Fig~\ref{fig5} shows the average VDEP values for each team. For example, Yokohama was able to defend with a high probability of recovering the ball, suggesting the probability of a high number of goals (see \nameref{S1_Data}). On the other hand, the probability of being attacked was also high, suggesting that the team adopted a high-risk, yet high-return, defensive tactic. Meanwhile, Hiroshima that had the fewest number of goals conceded in the league (see \nameref{S1_Data}), shows high probability of ball recovery and low probability of being attacked, suggesting that these properties led to the small number of goals conceded.

Statistically, we performed the correlation analysis between the team's performance over the whole season and the evaluation indices (the data is shown in \nameref{S1_Data}).
$R_{vdep}$ had moderate positive correlations with winning points ($r_{16} = 0.397, p = 0.103$), and low correlation with goals scored ($r_{16} = 0.342, p = 0.162$) and goals conceded ($r_{16} = -0.291, p = 0.239$) .
Meanwhile, $S_{vaep}$ had moderate positive correlation with goals scored ($r_{16} = 0.497, p = 0.034$), but slight almost negligible relationships with winning points ($r_{16} = 0.177, p > 0.05$) and goals conceded ($r_{16} = -0.098, p > 0.05$).
In the case of VDEP, the correlation coefficients with the game performances and those with the entire season were similar, whereas, in VAEP, the associations differed. 

\section*{Discussion}
In this study, we proposed a method to comprehensively evaluate a team's defense related to the team's performance, based on the prediction of ball recovery and being attacked (which occur more frequently than goals), using player actions and positional data of all players and the ball.
First, we verified the proposed and existing indices based on the prediction performance. Second, we quantitatively analyzed the defensive evaluations of the proposed and existing methods. 
Finally, we discuss the limitations of the proposed methods and future perspectives.

VAEP \cite{Decroos19} and \textcolor{black}{the proposed VDEP} evaluate players and teams based on the assumption of better prediction performance. 
To validate the classifiers, the previous study \cite{Decroos19} used AUC and Brier scores. 
However, as mentioned above, these indices may not be reliably evaluated because they include a large number of true negatives.
Therefore, we computed the F1 score and the results showed that the VDEP method predicted true positives correctly, while the VAEP did not.
This suggests that the VDEP method was a reliable method that can evaluate defensive performances based on better predictions.

Regarding the team evaluations using the proposed and existing indices, the correlation analysis revealed a moderate positive correlation between the season outcome (winning points) and the proposed VDEP value, whereas there were strong positive correlations between the game outcome (winning points and goals scored) and the existing VAEP value \cite{Decroos19}.  
Furthermore, overall, in the VDEP value, the correlation coefficients with the analyzed game performances and those with the entire season were similar, whereas those of the VAEP value were very different.
These results suggest that $R_{vdep}$ could be a well-balanced indicator to evaluate both attacks (after the ball recovery) and the defense itself (prevention of being attacked and the ball recovery).
On the other hand, the VAEP method \cite{Decroos19} is based on the prediction of offensive play and shows no correlation with the goals conceded. 
We expect that the use of VDEP in addition to the various indicators used so far will lead to the continuous strengthening of the team, regardless of immediate wins and losses which would be associated with contingent factors.

There are several recommendations for future studies. 
The first is the increase in the number of analyzed games for better prediction of longer-term game performances.
The second is the determination of the weighting constant $C$ in Equation \ref{eq:vvdep} for ball recovery and being attacked.
Although this study determined $C$ based on the number of occurrences of both events, the constant should be determined in more suitable ways for the practical values in soccer.
The third is an evaluation of the proposed method. Since such a new evaluation index often has no ground truth (or golden standard), we cannot validate our method using ground truth. In our future work, we need to construct a quantitative framework to evaluate such a method or to perform a subjective evaluation.
The last is the evaluations of individual players. 
Since VDEP evaluates team defense, it is difficult to evaluate the performance of individuals. For example, future studies are necessary to compute the change in VDEP when a player moves in different directions.

\section*{Acknowledgments}
We would like to thank Atom Scott and Masaki Onishi for their valuable comments on this work.

\section*{Supporting information}


\paragraph*{S1 Data.}
\label{S1_Data}
{\bf Analyzed data.}

\paragraph*{S1 Text.}
\label{S1_Text}
{\bf General description about 19 events in this study.}

\paragraph*{S1 Fig.}
\label{S1_Fig}
{\bf Confusion matrices of four classifiers.} The numbers of actual and predicted pass recoveries in VDEP, being attacked in VDEP, scores in VAEP, and concedes in VAEP are shown. Note that these are the results of test data in the last 9 games (other games were used for training). 

\bibliography{main}


\end{document}